%% file: main.tex
\documentclass[runningheads]{llncs}
\usepackage[T1]{fontenc}
\usepackage{graphicx}
\usepackage{amsmath}
\usepackage{amssymb}
\usepackage{booktabs}
\usepackage{xcolor}

\begin{document}
\title{Leveraging Cardiac Imaging to Improve ECG-Based Detection of Chagas Disease in Resource-Constrained Settings}

\titlerunning{ECG Detection of Chagas Disease in RCS}
%
%

\author{
Laura Alvarez-Florez\inst{1,2,3}$^\textbf{*}$ 
Daniel Uyterlinde\inst{3}$^\textbf{*}$ \and
Samuel Ruipérez-Campillo\inst{5} \and
Lukas P. A. Arts\inst{3,4} \and
Folkert W. Asselbergs\inst{3,4} \and
Fleur V. Y. Tjong\inst{3,4}
}

\authorrunning{L. Alvarez-Florez \& D. Uyterlinde et al.}

\institute{
Department of Biomedical Engineering and Physics,
Amsterdam University Medical Center,
The Netherlands
\and
Quantitative Healthcare Analysis Group,
University of Amsterdam,
The Netherlands
\and
Department of Clinical and Experimental Cardiology,
Amsterdam University Medical Center,
The Netherlands
\and
Amsterdam Cardiovascular Sciences,
Amsterdam University Medical Center,
The Netherlands
\and
Department of Computer Science,
ETH Zurich,
Switzerland
\\
\textbf{$^*$ These authors contributed equally to this work.}\\
\text{Corresponding author:} l.alvarezflorez@amsterdamumc.nl
}


%
\maketitle              
%


\begin{abstract}
Chagas disease is a major cause of cardiomyopathy in Latin America. Cardiac magnetic resonance (CMR) imaging can characterize its structural abnormalities, but scanners and expert readers remain scarce in endemic regions. Electrocardiography (ECG) is inexpensive and widely available, yet structural disease must be inferred indirectly from electrical signals. We propose to transfer CMR-derived structural knowledge to ECG through contrastive pre-training. Using 63,193 paired ECG–CMR examinations from the UK Biobank, we align an ECG encoder with a clinically grounded CMR embedding space using an asymmetric InfoNCE objective. Despite seeing no Chagas cases during pre-training, the resulting representation improves ECG-based Chagas detection. Across CODE-15\% and SaMi-Trop, a frozen linear probe achieves an AUROC of 0.851 and sensitivity at the top 5\% of predicted risk (Top5\%-TPR) of 0.427 in five-fold cross-validation, compared with 0.827 and 0.377 for an unaligned ECG-FM baseline. On the PhysioNet/CinC 2025 Challenge test set, our model obtains the highest AUROC on SaMi-Trop-3 and the best ELSA-Brasil challenge score among the three top-performing methods, indicating that imaging-supervised ECG representations can generalize to populations and settings beyond the pre-training distribution.

\end{abstract}

\input{01_introduction}

\input{02_related-work}

\input{03_methodology}
\input{04-experiments}

\input{05-conclusion}

\bibliographystyle{splncs04}
\bibliography{references}

\end{document}

%% file: 01_introduction.tex
\begin{figure}[t]
\centering
    \includegraphics[width=\textwidth]{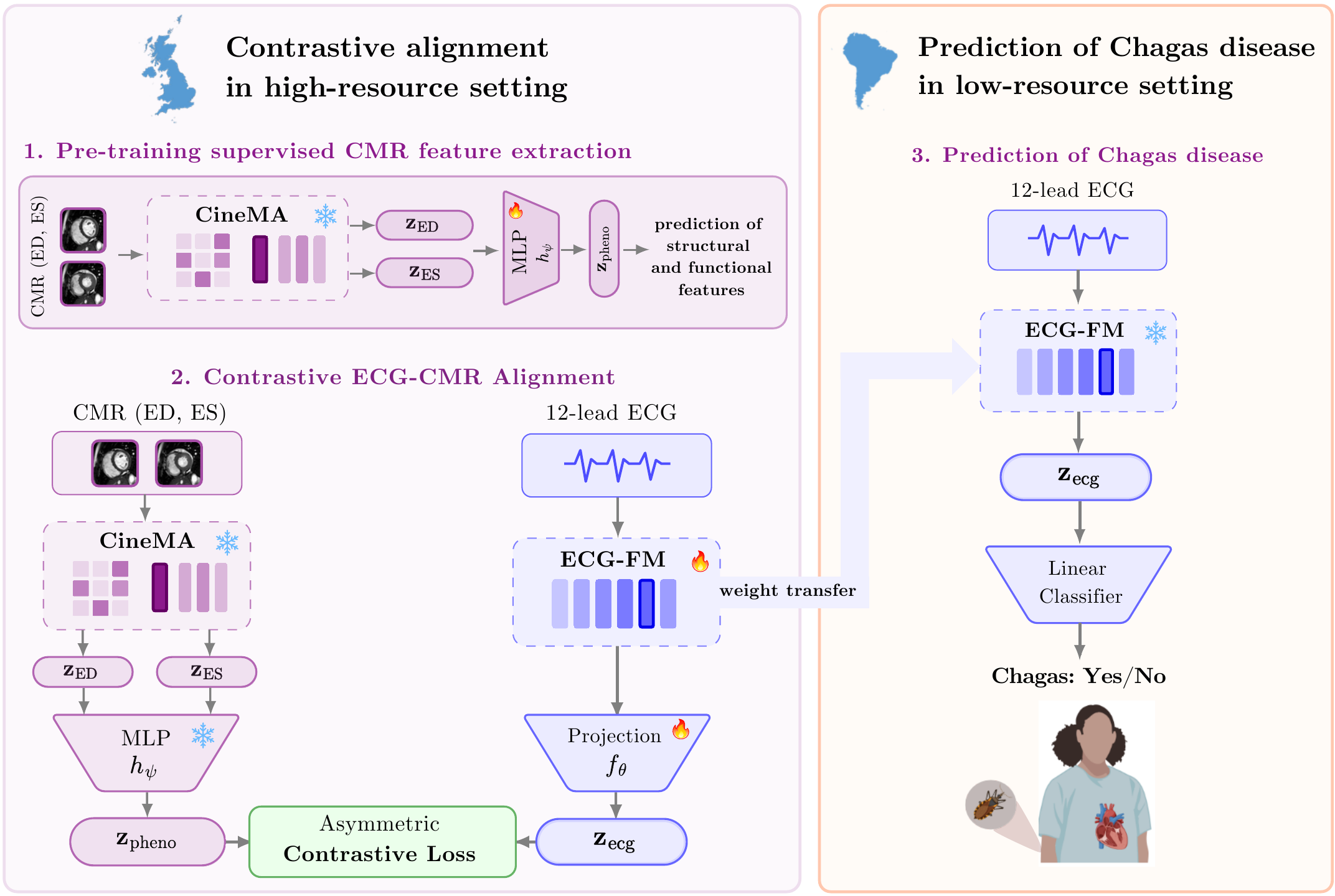} 
    \caption{Overview of the proposed method for ECG prediction of Chagas disease.}
    \label{fig:method}
\end{figure}

\section{Introduction}

Cardiac magnetic resonance (CMR) imaging is widely regarded as the reference standard for assessing cardiac structure and function and for phenotyping cardiovascular disease~\cite{pennell2010cardiovascular}. Yet the cost, infrastructure, and specialist expertise it demands place it out of reach for routine use in much of the world: Latin America, South-East Asia, and Africa have roughly 4-, 15-, and up to 30-fold fewer MRI scanners per capita respectively compared to European countries~\cite{brady2025current,hilabi2023impact}. The 12-lead electrocardiogram (ECG), by contrast, is inexpensive, portable, and available across nearly all healthcare settings. As a result, much of the world's cardiovascular disease burden, particularly in low- and middle-income countries (LMICs), is assessed with a modality that records the heart's electrical activity but reflects its underlying structure and function only indirectly.

This disparity in cardiac assessment is consequential for Chagas cardiomyopathy. Caused by \textit{Trypanosoma cruzi} infection, Chagas disease is a neglected tropical disease and a leading cause of non-ischemic cardiomyopathy in Latin America \cite{ECHAVARRIA2021100507}. Its manifestation is fundamentally structural, characterized by myocardial fibrosis, ventricular remodeling, and progressive ventricular dysfunction  \cite{ECHAVARRIA2021100507}. These abnormalities can be directly visualized with CMR, yet, access to the imaging modalities capable of directly characterizing these structural abnormalities remains limited in many affected and endemic regions. 

One promising approach to narrowing this gap is multimodal representation learning, where paired ECG and CMR examinations are used to teach ECG models about cardiac structure and function without requiring imaging at deployment. Several studies have demonstrated that contrastive alignment between ECG and CMR representations yields ECG encoders that outperform ECG-only self-supervised approaches on downstream cardiovascular prediction tasks \cite{ding2024cross,turgut2025ecgcmr,alvarezflorez2026dualphasecrossmodalcontrastivelearning,ding2026generating,selivanov2026PTACL}. By learning from paired ECG and imaging, these approaches encourage ECG representations to capture aspects of cardiac structure and function that are otherwise only directly observable through imaging. However, their evaluation has largely been restricted to populations and disease distributions similar to those used during model development, predominantly in European and North American cohorts.  It therefore remains unknown whether the structural representations transfer under domain shift, and in particular, whether they retain diagnostic value for diseases whose patients and imaging are entirely absent from the pre-training data.

In this work, we propose ECG-CMR contrastive pre-training as a mechanism for transferring structural cardiac knowledge from imaging populations to ECG-based diagnosis in imaging-scarce settings. We pre-train an ECG encoder by contrastive alignment against paired ECG-CMR examinations from the UK Biobank~\cite{ukbb}, and subsequently fine-tune the resulting representation on Chagas disease detection. Because CMR is required only during pre-training, deployment relies exclusively on ECG, preserving applicability in resource-constrained settings. Importantly, Chagas disease is absent from the paired ECG-CMR cohorts used in the training data, providing a stringent test of whether structural cardiac information learned from imaging can be transferred through ECG representations to a distinct disease never observed during learning.

We make three contributions.  First, we propose an imaging-supervised ECG pre-training framework that distills structural cardiac information from paired ECG-CMR data into deployable ECG representations. Second, we demonstrate that this structural information transfers to Chagas disease detection despite the complete absence of Chagas patients and Chagas-related imaging during pre-training. Third, we validate the approach on the PhysioNet/Computing in Cardiology 2025 Chagas Disease Detection Challenge benchmark~\cite{chagaschallenge}, achieving competitive performance while outperforming strong ECG-only baselines. Together, our findings suggest that structural cardiac knowledge acquired from imaging cohorts can be distilled into ECG representations and transferred to other diseases, populations, and healthcare settings beyond those observed during pre-training, providing a pathway toward more equitable and accessible cardiovascular diagnoses.

%% file: 02_related-work.tex
\section{Related Work}

\subsection{Deep Learning for ECG and Chagas Cardiomyopathy} 

Self-supervised learning and large-scale foundation models have substantially advanced ECG representation learning by learning generalizable features that can be transferred across a wide range of downstream cardiovascular tasks~\cite{mehari2021sslecg,li2025ecgfounder}. Driven in part by initiatives such as the PhysioNet Challenge,  deep learning methods have been extended to the prediction of Chagas cardiomyopathy from ECG. These efforts, focused on convolutional neural networks and Transformers ~\cite{jidling2023screening,erlacher2025swissbeatsnet}, are trained on disease-specific cohorts or fine-tuned from models pre-trained on large ECG datasets (e.g., PTB-XL \cite{ptbxl} or CODE-15\% ~\cite{ribeiro2021code}), thus remaining strictly unimodal. The models must therefore infer the complex structural patters associated with Chagas disease purely from electrical signals, without structural supervision during training.

\subsection{Multimodal ECG-CMR Representation Learning} 
To bridge the gap between electrical signals and cardiac mechanics, recent works have proposed multimodal contrastive learning to align ECG signals with CMR images. Methods such as MMCL \cite{turgut2025ecgcmr}, PTACL \cite{selivanov2026PTACL}, and ECCL \cite{ding2024cross} utilize large-scale paired datasets (predominantly the UK Biobank \cite{ukbb}) to project ECGs into a shared latent space with CMR embeddings. At inference, the ECG encoder can be deployed unimodally, retaining structural representations acquired during pre-training. 
While these models achieve state-of-the-art performance from ECG on structural phenotype regression, their evaluations are strictly confined to the populations and pathology distributions of their pre-training cohorts and do not assess the transferability of these representations beyond the training data.

%% file: 03_methodology.tex
\section{Methodology}
\label{sec:methodology}

Our framework operates in two phases (Fig \ref{fig:method}). In the pre-training phase, we use paired ECG and CMR examinations from the UK Biobank~\cite{ukbb} to align an ECG encoder with a clinically grounded structural target space via contrastive learning. In the deployment phase, the structurally-aware ECG encoder is deployed unimodally and adapted for Chagas disease classification in an unseen, imaging-scarce cohort.
 
\subsection{Unimodal ECG and CMR Encoders}
Rather than learning cardiac features from scratch, we initialize our framework with high-capacity unimodal foundation models to isolate the effect of cross-modal transfer. For the ECG modality, we use ECG-FM~\cite{mckeen2025ecgfm}, a hybrid CNN-Transformer architecture pre-trained via masked autoencoding and contrastive learning on 1.4 million 12-lead ECGs. For CMR, we use CineMA~\cite{fu2025cinema}, a multi-view convolution-transformer masked autoencoder pre-trained on cine CMR sequences that maps a single timeframe of a CMR volume of short-axis and (2-, 3-, and 4-chamber) long-axis views into a consistent spatial representation. During cross-modal alignment, the CMR encoder remains strictly frozen, both for computational efficiency and to provide a stable target.
 
\subsection{Clinically Grounded Target Space}
For ECG-CMR alignment, the target representation must encode not only inter-subject anatomical variation but also phase-dependent cardiac function. The raw CineMA embedding space, however, is near-degenerate: end-diastolic and end-systolic representations are nearly indistinguishable (average end-diastole to end-systole cosine similarity $>0.99$), so static anatomy dominates the variance and obscures subtle functional dynamics.
 
To construct a well-conditioned target that balances anatomy and function, we extract the end-diastolic ($\mathbf{z}_{\text{ED}}$) and end-systolic ($\mathbf{z}_{\text{ES}}$) representations from the frozen CineMA encoder. Rather than aligning to these raw vectors, we forward their concatenation through a supervised multi-layer perceptron (MLP):
\begin{equation}
    \mathbf{z}_{\text{pheno}}^{(i)} = h_{\psi}\left(\left[\mathbf{z}_{\text{ED}}^{(i)} ; \mathbf{z}_{\text{ES}}^{(i)}\right]\right) \in \mathbb{R}^{256}.
\end{equation}
The MLP $h_{\psi}$ is supervised to regress a set of ground-truth structural and functional phenotypes: left ventricle ejection fraction (LVEF), left ventricle end-diastolic volume (LVEDV), LV mass, right ventricle end-diastolic volume (RVEDV), left ventricle global longitudinal strain (LVGLS), Atrial Fibrillation (AFib), myocardial infarction (MI). Once trained, $h_{\psi}$ is frozen. This creates a $256$-dimensional, clinically grounded bottleneck ($\mathbf{z}_{\text{pheno}}$) that concentrates representational capacity on the dimensions of CMR space most relevant to cardiovascular pathology, stripping away redundant anatomical noise.
 
\subsection{Asymmetric Contrastive Pre-training}
To transfer this structural knowledge to the ECG, we employ an asymmetric contrastive alignment objective. Given a batch of $B$ patient pairs, let $\mathbf{z}_{\text{ecg}}^{(i)} = g_{\phi}(f_{\theta}(\mathbf{x}_{\text{ecg}}^{(i)}))$ denote the $L_2$-normalized ECG projection extracted from the optimal intermediate layer of the ECG-FM encoder (layer 9, found by linear probing every intermediate layer against several cardiac phenotypes), and let $\mathbf{z}_{\text{pheno}}^{(i)}$ denote the corresponding $L_2$-normalized CMR target.
 
We optimize an asymmetric InfoNCE loss, where gradients flow exclusively through the ECG encoder $f_{\theta}$ and its projection head $g_{\phi}$:
\begin{equation}
    \mathcal{L}_{\text{CL}} = -\frac{1}{B}\sum_{i=1}^{B} \log \frac{\exp(\mathbf{z}_{\text{ecg}}^{(i)} \cdot \mathbf{z}_{\text{pheno}}^{(i)} / \tau)}{\sum_{k=1}^{B} \exp(\mathbf{z}_{\text{ecg}}^{(i)} \cdot \mathbf{z}_{\text{pheno}}^{(k)} / \tau)},
\end{equation}
where $\tau=0.1$ is the temperature parameter. By holding the target space strictly fixed, the asymmetric InfoNCE objective forces the ECG embeddings to organize themselves according to the clinical geometry of the CMR space, preventing the contrastive update from distorting the underlying structural manifold.
 
\subsection{Downstream Unimodal Adaptation}
Following pre-training, the CMR branch is discarded. The aligned ECG encoder $f_\theta$ is deployed unimodally on the Chagas disease dataset (CODE-15\% and SaMi-Trop). We evaluate it as a frozen feature extractor with a linear classification head trained on Chagas labels. Because neither Chagas patients nor Chagas-specific structural remodeling were present in the UK Biobank pre-training data, performance on this task directly measures the zero-shot generalizability of the learned structural inductive biases to a neglected cardiovascular domain.

%% file: 04-experiments.tex
\section{Experiments}
\subsection{Datasets} 

We pretrain our proposed method on the UK Biobank~\cite{ukbb}. The UK Biobank provides 63,193 matched 12-lead ECG and multi-view cine CMR volumes. We use 47,683 for alignment training, and 15,510 for validation and held-out evaluation, stratified by LVEF, LV mass, sex and AFib diagnosis. UK Biobank ECGs are 10-second 12-lead recordings sampled at 500 Hz; each is split into two 5-second segments compatible with ECG-FM input. During training, either of these segments is stochastically sampled as data augmentation. CMR sequences provide short-axis and long-axis (2-, 3-, 4-chamber) cine views. 

We evaluate on the training data released for the PhysioNet 2025~\cite{chagaschallenge}. The combined datasets of CODE-15\%~\cite{ribeiro2021code} and SaMi-Trop~\cite{cardoso2016longitudinal} yields approximately 345,000 ECGs with 8,192 Chagas-positive cases ($\sim$2.4\% prevalence). For the challenge-protocol comparison we additionally include PTB-XL~\cite{ptbxl} as additional Chagas-negative training data (21,801 ECGs), matching the evaluation pool of van Santvliet et al. \cite{team-biomed}.

\subsection{Experimental Setup}
 
For the ablation study, we first compare our proposed model against \textbf{ECG-FM (no alignment)}, which uses the frozen ECG-FM encoder without cross-modal alignment, establishing the performance of ECG-only self-supervised pre-training. We further compare our method with \textbf{Jidling et al.}~\cite{jidling2023screening}, who trained an ensemble of fifteen 1D-ResNet models end-to-end using the full CODE dataset together with SaMi-Trop. Finally, we compare against \textbf{van Santvliet et al.}~\cite{team-biomed}, the winning approach in the PhysioNet 2025 Chagas Challenge, which fine-tuned a ViT-based ECG foundation model on CODE-15\%, SaMi-Trop, and PTB-XL.

\subsection{Implementation Details}
ECG-FM is fine-tuned with AdamW ($\eta_{\mathrm{enc}}=10^{-5}$, 
$\eta_{\mathrm{head}}=10^{-3}$, weight decay $10^{-3}$), batch size 256 on a single NVIDIA A100, with cosine annealing and early stopping on held-out linear probing scores. For downstream Chagas prediction, layer 9 of the aligned encoder serves as a frozen feature extractor with a linear head trained in five-fold cross-validation. Layer 9 was selected independently of Chagas labels based on linear-probe performance across clinically relevant cardiac phenotypes. Class imbalance is addressed via positive-class reweighting and checkpoint selection uses AUPRC. The challenge submission ensembles this linear probe 
with a full end-to-end fine-tuning run initialized from the contrastive checkpoint ($\eta_{\mathrm{enc}} = 10^{-5}$, $\eta_{\mathrm{head}} = 10^{-4}$).



\section{Results}
In this section, we evaluate the value of performing cross modal alignment as a pretraining step, and compare our proposed method with existing state-of-the-art methods for prediction of Chagas disease.

\textbf{Value of cross-modal alignment.}
To evaluate the benefit of learning from imaging during the pretraining phase, we compared our proposed method with the baseline ECG-FM method before cross modal pretraining. The results reported on Table 1 show an increase of 5 points in Top5\%-TPR, showing that CMR-derived structural information, such as ventricular geometry or myocardial mass, transferred through contrastive alignment, provides genuine diagnostic value for Chagas cardiomyopathy detection, even though none of this structural information was observed in the context of Chagas disease during pre-training.

\textbf{Comparison with existing methods for Chagas prediction.} Table~\ref{tab:main} reports five-fold cross-validation on CODE-15\% and SaMi-Trop, enabling direct comparison to the Jidling et al.\ baseline. ECG-CMR contrastive pre-training outperforms the unaligned ECG-FM baseline by 5.0 percentage points in Top5\%-TPR and surpasses the AUROC of $0.80$ reported by Jidling et al., whose model was \emph{trained specifically on Chagas data}. Our encoder, by contrast, has never encountered a Chagas patient or Chagas-related CMR.
 
\begin{table}[h]
\vspace{-1.5em}
\centering
\caption{Five-fold cross-validation on CODE-15\% + SaMi-Trop (without PTB-XL). All models use a frozen encoder and linear probe. The Jidling et al.\ model was trained end-to-end on Chagas data; ours was not. Dash line indicates values not reported.}
\label{tab:main}
\begin{tabular}{lcc}
\toprule
Method & Top5\%-TPR & AUROC \\
\midrule
Jidling et al.~\cite{jidling2023screening} & --- & 0.800 \\
ECG-FM (no alignment) & $0.377 \pm 0.014$ & $0.827$ \\
\textbf{Ours (CMR-aligned, linear probe)} & $\mathbf{0.427 \pm 0.022}$ & $\mathbf{0.851}$ \\
\bottomrule
\end{tabular}
\vspace{-1.5em}
\end{table}
 
\begin{table}[b!]
\vspace{-1em}
\centering
\caption{Five-fold cross-validation on CODE-15\% + SaMi-Trop + PTB-XL (challenge protocol). Both our method and the Van Santvliet et al.\ frozen ablation use a frozen encoder and linear probe. Dash line indicates values not reported.}
\vspace{-0.5em}
\label{tab:challenge}
\begin{tabular}{lcc}
\toprule
Method & Top5\%-TPR & AUROC \\
\midrule
Van Santvliet et al.~\cite{team-biomed} (linear probe) & $0.381 \pm 0.003$ & --- \\
\textbf{Ours (CMR-aligned, linear probe)} & $\mathbf{0.431 \pm 0.005}$ & $\mathbf{0.852 \pm 0.007}$  \\
\bottomrule
\end{tabular}
\end{table}

Table~\ref{tab:challenge} reports results under the full challenge protocol. Our frozen linear probe outperforms the challenge winner's frozen-backbone ablation by 5.0 percentage points in Top5\%-TPR; the remaining gap to their end-to-end fine-tuned result reflects the frozen-probe constraint rather than a difference in representation quality. Table~\ref{tab:official} reports our official submission on the hidden test sets. Our overall Challenge score (0.269) places just below the top three teams (0.280--0.323). Two results stand out. First, on the SaMi-Trop-3 dataset we attain the highest AUROC among compared methods (0.773 vs.\ 0.767 for the winner). Second, on ELSA-Brasil, the hardest cohort owing to demographic and acquisition differences from the training pool, our submission achieves the best Challenge score (0.132).

 
\begin{table}[h]\centering\small\setlength{\tabcolsep}
{4pt}
\caption{Official PhysioNet/CinC 2025 Chagas Challenge test-set results (challenge score / AUROC per cohort). Overall is the mean challenge score across the three test cohorts, matching the official leaderboard aggregation. Ranked teams had access to the REDS-II validation set for threshold calibration; we did not.}
\vspace{-0.5em}
\label{tab:official}
\resizebox{\textwidth}{!}{%
\begin{tabular}{lcccc}
\toprule
Method & REDS-II & SaMi-Trop-3 & ELSA-Brasil & Overall \\
\midrule
1st, Van Santvliet et al.~\cite{team-biomed} & \textbf{0.468} / \textbf{0.777} & \textbf{0.376} / 0.767 & 0.125 / 0.566 & \textbf{0.323} \\
2nd, Nicolson et al. & 0.357 / 0.735 & 0.375 / 0.739 & 0.118 / 0.567 & 0.283 \\
3rd, Hong et al. & 0.382 / 0.704 & 0.329 / 0.749 & 0.129 / \textbf{0.626} & 0.280 \\
Ours & 0.350 / 0.739 & 0.326 / \textbf{0.773} & \textbf{0.132} / 0.556 & 0.269 \\
\bottomrule
\end{tabular}}
\vspace{-1em}
\end{table}

%% file: 05-conclusion.tex
\section{Discussion}
We presented an imaging-supervised ECG pre-training framework for Chagas disease detection. By aligning ECG representations with a clinically grounded CMR target space on the UK Biobank, we distill structural cardiac knowledge into a deployable ECG encoder without needing access to CMR during inference.

Our results show that contrastive ECG-CMR pre-training provides diagnostic benefit beyond the diseases and populations present in the imaging cohort. By comparing against an unaligned ECG baseline, we show that structural cardiac knowledge distilled from paired CMR transfers to Chagas cardiomyopathy, despite Chagas-specific imaging manifestations not being explicitly represented during pre-training. The results suggest that the model captures general structural and functional cardiac characteristics that remain informative for Chagas disease across populations. These results could be extrapolated in future work to other cardiac conditions where imaging is unavailable, suggesting a broader pathway for improving ECG-based diagnosis of neglected diseases in resource-constrained settings.

We have also positioned our method within the current landscape of Chagas disease prediction. The CMR-aligned encoder outperforms both ECG-only self-supervised baselines and models trained end-to-end directly on Chagas-labeled data, establishing a new reference point for what is achievable without any disease-specific supervision. On the official challenge benchmark, while overall performance was just below top-ranked, results on the ELSA-Brasil cohort outperform those of the top three challenge winners. As the dataset representing the greatest demographic and acquisition shift from the training distribution, this improvement suggests that structural features learned from imaging transfer most reliably to populations and settings furthest from the pre-training data. Future work could build on these findings by incorporating additional cardiac imaging modalities during pre-training to further enrich the structural and functional information transferred to the ECG encoder. Moreover, the present analysis does not identify which specific structural phenotypes contribute to Chagas prediction. Investigating the relationship between the learned ECG representations and Chagas-relevant cardiac manifestations would provide further insight into the mechanisms underlying this cross-disease transfer.

From a translational perspective, the proposed framework separates the resource intensive multimodal pre-training stage from downstream application. While developing the representation requires access to large paired ECG--CMR datasets and substantial computational resources, these requirements are limited to pre-training. Once trained, the CMR branch is discarded and the resulting encoder operates using only a standard 12-lead ECG. This creates the possibility of learning multimodal cardiac representations in data-rich settings and subsequently transferring them to populations and healthcare settings where advanced cardiac imaging is less accessible. Future work should evaluate this transfer prospectively across Latin American populations and assess the computational requirements for deployment in resource-constrained settings.

\section{Impact in RCS} By distilling structural cardiac knowledge into an ECG encoder, this framework enables diagnosis in settings where advanced imaging infrastructure is not available.